# Intelligent Unmanned Explorer for Deep Space Exploration

**T. Kubota** and **T. Yoshimitsu**

Institute of Space and Astronautical Science
Japan Aerospace Exploration Agency
Sahamihara, Japan.
e-mail: kubota@isas.jaxa.jp

**Abstract**

In recent years, such small body exploration missions as asteroids or comets have received remarkable attention in the world. In small body explorations, especially, detailed in-situ surface exploration by tiny rover is one of effective and fruitful means and is expected to make strong contributions towards scientific studies. JAXA/ISAS is promoting MUSES-C mission, which is the world's first sample and return attempt to/from the near earth asteroid. Hayabusa spacecraft in MUSES-C mission took the tiny rover, which was expected to perform the in-situ surface exploration by hopping. This paper describes the system design, mobility and intelligence of the developed unmanned explorer. This paper also presents the ground experimental results and the flight results.

## 1 Introduction

Small planetary bodies such as asteroids, comets and meteorites in deep space have received worldwide attentions in recent years. These studies have been motivated by a desire to shed light on the origin and evolution of the solar system. Hence, exploration missions for small bodies have been carried out continuously since the late 1990s. To date, the missions of NEAR [1], Deep Space 1 [2], Deep Impact [3], and Stardust [4] have been successfully performed, while MUSES-C[5] and Rosetta[6] are currently in operation. These missions have mainly provided remote sensing in the vicinity of the small body, at a distance which cannot be attained from the earth. In-situ observations of small bodies are scientifically very important because their sizes are too small to have experienced high internal pressures and temperatures, which means they should preserve the early chemistry of the solar system. For future missions, in-situ surface observation by robots will make strong contributions towards those studies.

In such deep space missions, ground based operation is very limited due to communication delays and low bit-rate communication. Therefore, autonomy is required for deep space exploration using rovers. On the other hand, because the gravity and surface terrain are not known in advance, robotics and artificial intelligence technology must be used for rovers to explore a minor body.

The Institute of Space and Astronautical Science (ISAS) of Japan launched the MUSES-C spacecraft toward Asteroid 1998SF36 in May 2003. After the launch, the spacecraft was renamed "HAYABUSA" which means falcon. The MUSES-C project is demonstrating key technologies required for future sample return missions from extra-terrestrial bodies as shown in **Figure 1**. The launch date was May 9th in 2003 and arrival at the asteroid was September 12th in 2005. Leaving the asteroid in April 2007, the spacecraft will return back to the earth in June 2010. HAYABUSA spacecraft has stayed for approximately three months around the asteroid and both mapping and sampling operations were carried out during that short period. For the MUSES-C mission, an asteroid surface exploration rover was developed, called MINERVA for MIcro/Nano Experimental Robot Vehicle for Asteroid. MINERVA is the first asteroid exploration rover to ever be developed and deployed. It is one of the technical challenges for the MUSES-C mission and its major objectives are as follows: establish a mobile system in the micro gravity environment of a small planetary body, demonstrate the autonomous exploration capability the rover is equipped with, and perform the first-ever planned scientific observations on an asteroid surface. On November 12th in 2005, the spacecraft Hayabusa deployed MINERVA at higher velocity than the escape velocity, and MINERVA could not arrive at the asteroid surface. However the spacecraft could communicate with MINERVA and it was confirmed that the health of MINERVA was very good and some autonomous functions were performed very well.

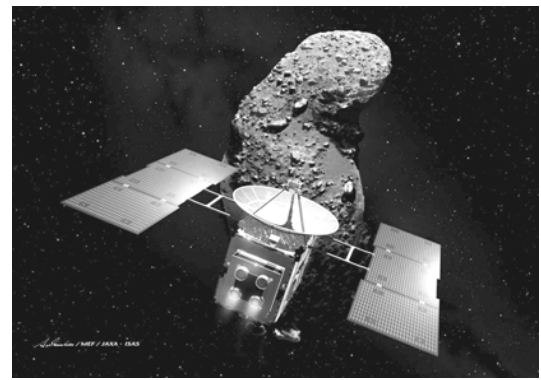

Figure 1:   MUSES-C mission

This paper presents the challenges of mobilizing a rover on a minor body surface and discusses the necessary intelligence such an exploration rover must have. Section 2 discusses possible mobility systems for the micro-gravity environment on a small body surface, and presents a hopping mechanism that was developed in MUSES-C mission. Section 3 describes the intelligent functions of the micro probe robot. Section 4 presents the asteroid exploration rover MINERVA and flight results, which has been designed to explore an asteroid surface for the MUSES-C mission. Conclusions are stated in Section 5.

## 2 Mobility system

The gravity on small asteroids such as Itokawa is on the order of 10 $\mu G$, which is extremely weak compared with that on Earth (1$G$ equals to one earth gravity). Naturally, the escape velocity from the surface is very small as well, on the order of 20 centimeters per second. The largest asteroid yet discovered, Ceres, has a diameter of more than 900 km, however the majority of asteroids have diameters less than several kilometers in size, and surface gravity lower than one milli-$G$. Rovers that are intended to move over a small body surface need a mobility system suited for this milli- or micro- gravitational environment.

### 2.1 Mobility under microgravity
The main principle to be considered is that the traction force which drives the robot horizontally is very small in a micro gravity environment. Traction force is obtained by the friction between the robot and the surface. Assuming that Coulomb's friction law holds true in a microgravity environment, the traction force $f$ is given by

$$f = \mu N \quad (1)$$

,where $\mu$ denotes the friction coefficient and $N$ is the contact force. In the case of a traditional wheeled mechanism, frequently used by planetary rovers on the Moon or on Mars, the contact force is opposite to the gravity, which results in the following equations,

$$f = \mu N = \mu mg \quad (2)$$

In Eq.(2), $m$ is the mass of the rover and $g$ is the gravitational acceleration on the surface. Eq.(2) indicates that the available friction for wheeled mobility is very small in the microgravity environment ( approximately $10^{-4}$ m/s$^2$ ). Thus, even if the friction coefficient is assumed to be large, friction between a robot and the surface becomes very low. Additionally, if the traction is larger than the maximum friction, the rover slips. Thus the available traction must be extremely small, which makes the horizontal speed extremely slow. Actually, NASA proposed a new robot [7] with wheeled mechanism as a payload of the MUSES-C mission, but it could have only driven only with the speed of 1.5 mm/s. To achieve distances of tens of meters, it would have used a mobility mechanism to hop.

Moreover, the surface of small bodies is a natural terrain with uneven surfaces. When the robot begins to move, any small disturbance may push the robot away from the surface. So the mobility system such as traditional wheeled mechanisms is not optimal under a micro gravity environment, where continuous robot-surface contact is not guaranteed. This reason motivates the search for another mobility system that specializes in hopping. Once a robot hops with some horizontal speed, it can move with no contact on the surface.

A hopping mechanism causes the robot to push the surface. This increases the contact force between the surface and the body artificially with the pushing force, which makes the friction larger and can provide mobility at a higher horizontal speed that cannot be attained by a wheeled mechanism. Once it has hopped into free space with some lateral speed, the robot moves in a ballistic orbit under the influence of the weak gravity. Lateral speed is also obtained by the friction acting during the short period after the actuator starts and before separation from the surface. In the hopping process, the friction force exerted by the hopping mechanism is expressed as:

$$f = \mu ( mg + F ) \quad (3)$$

where $F$ denotes the artificial pushing force which makes the robot hop. In the case $mg \ll F$, the friction force is approximated as:

$$f \sim \mu F \quad (4)$$

Thus, a robot can use a friction force independent on the value of the gravity acceleration. In order to press its mass to the ground, some mechanism is required. By increasing the pushing force, the hopping mechanism is able to attain higher speeds which cannot be attained in principle by a wheeled mechanism. The maximum allowed speed is the escape velocity from the surface. If a larger speed than the escape velocity is achieved, the robot may never return to the surface again.

### 2.2 Hopping mechanism
There are several ways [8-16] to make a robot hop. The hopping mechanisms are categorized into two groups. One is to use repulsed force by striking, sticking, or kicking the surface. The other is to use reaction force by rotating an internal torquer. The newly developed hopping mechanism is shown in **Figure 2**. MINERVA installed such a mechanism and incorporates a torquer inside of its structure. All the previous studied designs have had a moving part outside of the rover body. However, MINERVA has no apparent moving parts outside, which has a big advantage in reliability for long-term outer-space use. After hopping into free space, the robot moves ballistically and comes back to the surface again, so long as its launch speed is less than the escape velocity from the surface. After a few bounces on the surface, it will settle and be ready to hop again. This action can be repeated, allowing the robot to explore the surface of the small planetary body. This proposed mechanism has several significant advantages:

1. The actuator is not outside the rover. Since the actuator is sealed inside the body, no consideration is required for dust, which may exist on the small body surface.

2. The torquer can also be used for attitude control during the ballistic orbit.

3. The contact force between the surface and the body is increased with help of the artificial pushing force made by the torquer, which makes the friction larger and provides for mobility at larger horizontal speeds.

4. DC motors can be used as a torquer, which are easily controlled. The imposed torque must be adjustable in order to control the hopping speed. A DC motor driven by PWM is used to provide torque adjustability.

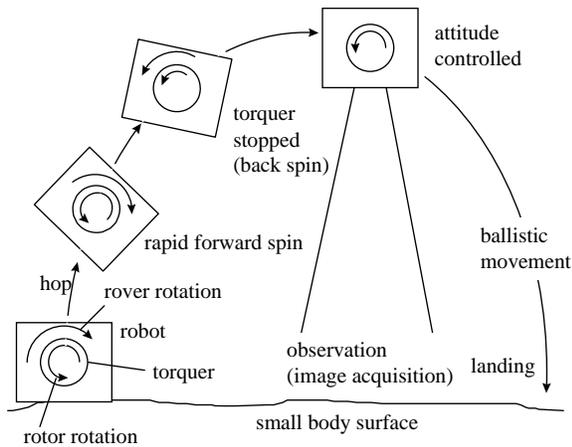

**Figure 2:** Hopping mechanism

### 2.3 Experiments
In order to confirm the hopping mechanism, microgravity experiments [17][18][19] were conducted at the Japan Microgravity Center (JAMIC), where 10 seconds of microgravity is obtained inside a capsule that free-falls 490 m. The experimental robot was developed and experimental results were compared with the simulation analyses. Five experiments were conducted, as shown in **Table 1**, each of which differs in the parameters of the motor-supplied PWM history. Video monitoring was performed for each experiment. For example, the sequence of images for experiment #5 is shown in **Figure 3** at intervals of 1/3 sec, where a PWM duty cycle of 50 % pulses was supplied for 0.5 sec.

In each experiment, the images from every 1/30 sec were extracted. The robot has many optical markers on its surface, whose three-dimensional coordinates in the robot-fixed coordinate system are known. The visible markers are tracked in each image and the robot position and attitude in the experimental apparatus are estimated in every image by using a least squares method. For example, the tracked visible markers are shown in **Figure 4** by squares for the image of experiment #5 after 8/3 sec had passed since the start of the torquer (corresponds to **Figure 3 (j)**).

**Table 1:** Specification of MINERVA

| case | duty ratio $d_m$ [%] | duration $t_m$ [sec] | voltage $V_P$ [V] |
|------|-----|-----|------|
| #1 | 100 | ∞ | 4.53 |
| #2 | 50 | ∞ | 3.54 |
| #3 | 25 | ∞ | 4.80 |
| #4 | 100 | 0.5 | 2.33 |
| #5 | 50 | 0.5 | 4.05 |

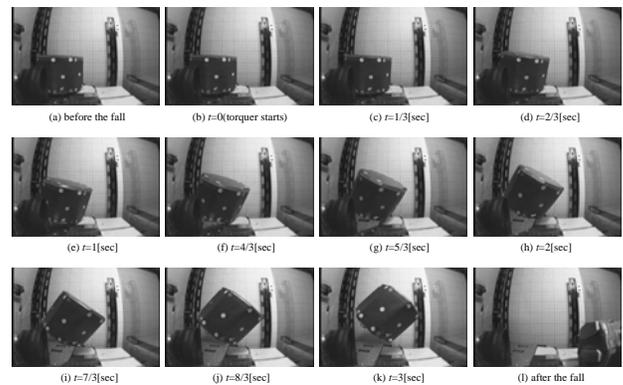

**Figure 3:** Experimental results

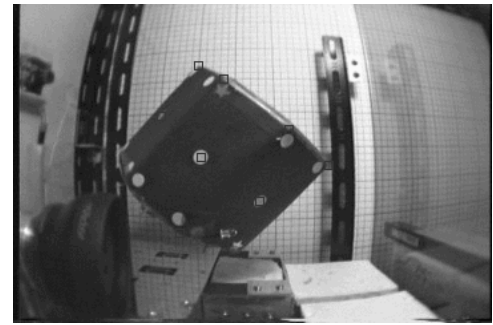

**Figure 4:** Tracking markers

### 2.4 Experimental study
After estimating the position of the robot in each image, the estimated positions are plotted with time. After the robot hops into free space, the robot goes straight as the gravity inside the apparatus is zero. The linear line is fitted on the plotted positions to derive the slope of the fitted line, which shows the estimated speed of the robot. For example, **Figure 5** shows the estimated positions of experiment #5, where the horizontal and vertical coordinates of the robot in the apparatus are plotted. The fitted linear lines are also overlaid in **Figure 5**, the slopes of which show the horizontal and vertical values of the hop speed. In this way, the hop speed and direction can be estimated.

under development but are not discussed further in this section.

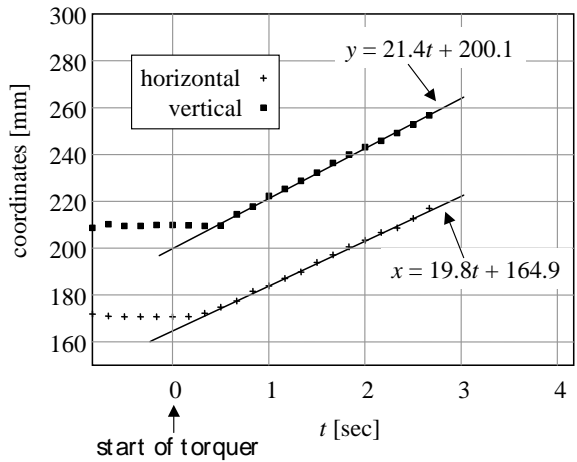

**Figure 5:** Experimental result (case #5)

**Table 2** summarizes all the experimental results of hop velocity and direction. The hop time is not shown in this table because it is difficult to distinguish whether or not the robot is contacting the surface. Using the hopping model described, simulation analyses were performed, which are shown in **Table 3**. The values used are shown in **Table 1**. The gravitational acceleration was set to be zero (gx=gy=0) and the coefficient of friction was measured and found to be approximately 1. The shape of the robot is a rectangle as viewed from the side and the distance from the center of the robot to the contact point is 156.2 mm, and the angle is equal to 219.8 deg.

**Tables 2** and **Table 3** compare the experimental results and simulation analyses respectively, and show consistency between the results. Of course both results will differ a bit because the simulation model is not completely correct, the experiments include a lot of errors, the adopted estimation method does not derive the true value of the hop speed, direction, etc. It can be concluded that the simulation method provides approximate estimates of the truth and the proposed strategy can deduce the hop speed. Based on these experiments, our conclusion is that the hopping control strategy is expected to work well on the surface of the small body.

With the proposed hopping strategy, the hop direction cannot be controlled because it is mainly dependent on the friction coefficient between the robot and the surface, which can neither be controlled nor estimated a priori. The hop speed is also affected by the friction coefficient, but here is not a major factor as the imposed torque largely dominates the hop speed. Thus the hopping strategy enables control of the hop speed. It may be difficult to control the hop direction, but by using the internal sensors and the acquired images of the surface of the planetary body, the hop speed and direction can be estimated during the ballistic phase, which enables the prediction of the friction coefficient and how much distance the robot flies. These techniques are

**Table 2: Experimental results**

| case | hop speed | | | hop direction $\theta_h$ [deg] |
|---|---|---|---|---|
| | horizontal $v_{hx}$ [mm/s] | vertical $v_{hy}$ [mm/s] | $v_h$ [mm/s] | |
| #1 | 50.3 | 47.2 | 69.0 | 46.8 |
| #2 | 38.7 | 32.6 | 50.6 | 49.9 |
| #3 | 32.0 | 26.4 | 41.4 | 50.5 |
| #4 | 19.8 | 20.5 | 28.5 | 44.0 |
| #5 | 19.8 | 21.4 | 29.1 | 42.8 |

**Table 3: Simulation results**

| case | hop speed | | | hop direction $\theta_h$ [deg] | hop time $t_h$ [sec] |
|---|---|---|---|---|---|
| | horizontal $v_{hx}$ [mm/s] | vertical $v_{hy}$ [mm/s] | $v_h$ [mm/s] | | |
| #1 | 60.5 | 50.6 | 78.8 | 50.1 | 0.70 |
| #2 | 35.5 | 29.9 | 46.4 | 49.8 | 1.14 |
| #3 | 28.2 | 23.9 | 36.9 | 49.7 | 1.40 |
| #4 | 24.8 | 23.8 | 34.4 | 46.2 | 0.50 |

## 3 Intelligence of unmanned explorer

In deep space missions, ground based operation is very limited due to communication delays and low bit-rate communication. Therefore, intelligence is required for small unmanned explorer.

### 3.1 Intelligent control system

The hierarchy of the onboard control system [20] is shown in **Figure 6**. The lowest levels close to the devices are conducted by independent controllers. The next level is the action layer controlling the actuators, sensors and communication devices. For example, "take the picture from camera (A)", "rotate the motor with duty pulse of X" are the commands on this level. The discrete commands to MINERVA from the earth can directly designate the commands on this layer when the CPU is working in the tele-operation mode.

The highest level is the behavior control layer, which calls the action level commands to compose autonomous actions. MINERVA has some operational modes [21], which differ from the program of this layer. When it is booted up, a certain Flash ROM area is first referred to in order to extract the operational mode and select the program. The switch of this operational mode is basically conducted by human command. The surface of the asteroid may be very cruel in temperature, which changes beyond the range of the onboard devices of MINERVA working well. The developed rover has the automatic capability to wake and shut down the CPU and is sensitive to the surrounding temperature so as to not work during hot or cold periods.

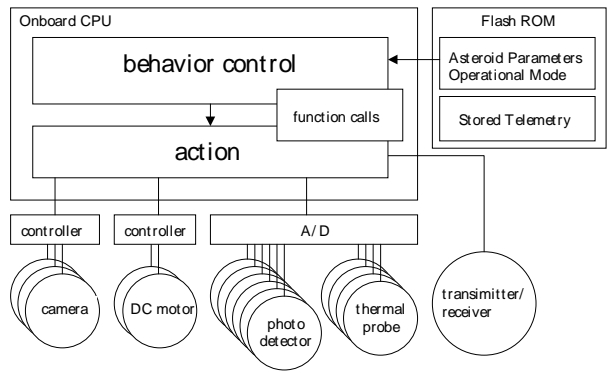

**Figure 6:** Intelligent control system

### 3.2 CPU system
A CPU with high performance and low power consumption is required for unmanned small body explorer. Based on radiation tests, the commercial CPU is guaranteed to work in the high radiation environment of outer space. However, compared with the latest CPU commercially available, for example, its processing speed is drastically slow with only 10 MIPS when the clock is driven at 10 MHz. The CPU system has small memory which preserves the program and small RAM as a working area. Additionally it should have a Flash ROM in which acquired data are stored. The surface of the asteroid may be a very harsh environment in terms of its temperature, which changes beyond the optimal operating range of the onboard devices. The rover has an automatic capability to wake up and shutdown the CPU depending on the surrounding temperature, so that operation is suspended during periods of extreme hot or cold.

### 3.3 Intelligent power supply system
The rover is powered by solar energy, supplemented by secondary batteries, which enable a few minutes of activity with no help from the sun when the solar arrays are blocked by shadows. The solar cells are attached on each face of the rover. The power generated is dependent on the attitude of the rover, with a maximum power expected when the rover is on an asteroid surface at a distance of 1 AU from the sun. Extra power is stored in the electric double layer capacitors. The onboard CPU works with the instantaneous power available from the solar arrays, but that is not sufficient for motor rotation, image capture and communication.

For such operations, additional power support from the charged capacitors is necessary. Unfortunately the capacitors gradually degrade when temperatures go higher than 130 deg C. After a couple of days on the asteroid surface, the capacitors can no longer be used. However even if the capacitors are fully degraded, as long as the solar energy is supplied, the computer and communication system works in a low consumption mode, but the hop performance is degraded.

### 3.4 Autonomous thermal control system
The operational temperature range of the onboard devices is from -50 deg C to +80 deg C. Beyond this range, the devices do not work well. The rover is equipped with an automatic capability to wake up and shutdown the onboard CPU depending on the surrounding temperature. When the temperature becomes more than +80 deg C or less than -50 deg C, the power supply to the CPU is suspended. The capacitors are not degraded if it is very cold, but will gradually degrade when the temperature goes higher than +130 deg C. After a couple of days on the asteroid surface, the capacitors can no longer be used because they will have experienced a couple of overheated daytimes. Even so, as long as the solar energy is supplied, the CPU works, but the hop performance is degraded and the rover may not move. The onboard software also monitors the internal temperature. If this comes toward the threshold of working temperature, the actions which consume much power are suspended and the data in RAM are transferred to the Flash ROM in order not to be lost by the sudden shutdown of the CPU system.

The onboard devices are covered by MLI heat insulation material. The heat capacity of the rover is very small as the temperature of the onboard devices is easily synchronized with the surrounding asteroid temperature. During the night, the asteroid temperature will fall below -100 deg C, but the temperature on the onboard devices will not drop as quickly because the heat transfer is blocked. This makes for a quick wake up of the rover at sunrise because the device temperature is not so low. Toward noon, the temperature may be rise higher than +80 deg C and the rover will take a nap for a while. A CPU with high performance and low power consumption.

### 3.5 Onboard image selection system
The taken pictures are compressed and evaluated by the onboard CPU. The images with the least amount of information, such as completely black images, are abandoned immediately. Images with greater information are saved in the Flash ROM, attached with its priority proportional to the amount of information. When the rover communicates with the spacecraft to send the acquired data, data with highest priority is transmitted first. Using this selection strategy, scientifically important images are transmitted effectively within a limited bandwidth.

### 3.6 Autonomous localization system
Localization to calculate the current attitude and position of the rover is needed for robust navigation. Estimation of absolute position on the asteroid can only be attained if the attitude of the rover is known. However MINERVA can not directly sense its attitude. So localization is a very difficult problem and onboard localization is only an optional function. If the rover were equipped with gyroscopes, once the attitude of the rover is calculated in one moment by some means, propagation would enable one to find the attitude at a future time. For MINERVA, however, a localization technique [22] is introduced for a rover not equipped with gyroscopes. After the rover stops in the evening and before the rover begins to move in the morning, the position and the attitude are reserved, when the solar directions are observed for a couple of times. The attitude of the rover with respect to inertial space can be estimated using the solar observation. Then the gravitational directions, which are estimated during the hop, as well as the hours of sunset and sunrise, are used to derive the absolute position and attitude in the asteroid-fixed frame.

### 4  Asteroid Exploration Rover

#### 4.1 MINERVA

ISAS has developed an asteroid rover MINERVA [23] which is an experimental payload of the HAYABUSA spacecraft. **Figure 7** shows the flight model of MINERVA. MINERVA had planned to explore the asteroid surface after being deployed from the HAYABUSA spacecraft. Its shape is a hexadecagon-cylinder with a diameter of 12 cm and a height of 10 cm, its weight was 591g. **Table 4** shows the specification of MINERVA. It is powered by solar energy, supplemented by condensers which enable a few minutes' worth of power when the sunlight is blocked by shadows. During cruise, MINERVA is also supplied with power by wired line from the OME(Onboard Mounted Equipment).

OME holds MINERVA during cruise. When MINERVA-deployment is triggered, the cover of OME is thrown away to allow MINERVA to be pushed towards the surface. After deployment, OME plays the role of a communication relay between MINERVA and the HAYABUSA spacecraft data recorder. Telemetry from MINERVA is received by OME via the antenna mounted on the asteroid-oriented face of the spacecraft. Commands from Earth are stored in the OME, and when communication becomes possible with MINERVA, the stored commands are transmitted. The communication bandwidth is 9,600 bps within a distance of 20 km.

#### 4.2 Onboard sensors

The appearance of MINERVA and the sensor allocations are shown in **Figure 8**. MINERVA has three CCD cameras mounted on the turntable. The camera sight direction can be controlled by rotating the turntable. There are camera windows at the center of the side faces of the rover, through which the cameras view the outside. All the cameras are commercially available and are sensitive at visible wavelengths. They have also passed radiation level tests and have been slightly tuned considering the hottest operational temperatures expected. The focal length cannot be adjusted onboard, thus the two cameras are set to look very closely and another one is to look far away.

MINERVA has six photo detectors (PDs) used as sun sensors. A single sun sensor senses the intensity of the incoming light, not the solar direction. Thus by using two outputs from the directly illuminated sensors, the solar direction can be detected. The allocation of the sun sensors is offset 90 deg from each other. At least three detectors will sense the light simultaneously and, by comparing the outputs of these detectors, the solar direction is computed. There are also pins sticking out from the body to protect the solar panels. Four of the pins are also used as thermal detectors, by which the temperature of the surface is measured.

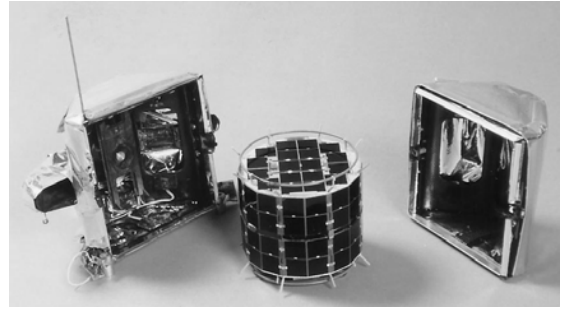

**Figure 7:**   Flight model of MINERVA

**Table 4:**   Specification of MINERVA

| Body size | Hexadecagonal cylinder $\phi$: 120[mm], height: 100[mm] |
|---|---|
| Weight | 591[g] |
| Mobility system | Hopping ability [9cm/s(max)] |
| Power supply | Solar cells (2.2[W] at 1[AU]) Condenser (4.6[V], 20[F]) |
| Onboard CPU | 32bit CPU(10M[Hz]) |
| Memory | ROM:512kB, RAM:2MB, Flash ROM:2MB |
| Temperature Range | Min:-50[deg C], Max:+80[deg C] |
| Communication | 9600bps (max range: 20[km]) |
| Payloads | CCD camera $\times$ 3, Sun sensor $\times$ 6, Thermometers $\times$ 6 |
| Power Consumption | 2.6[W] for actuators (max.), 1.8[W] for communication 8[W] for camera, 0.8[W] for on-board computer |

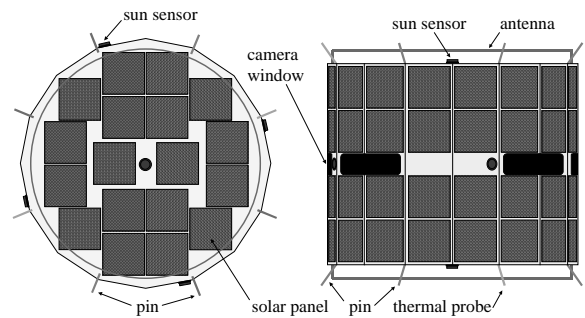

**Figure 8:**   Onboard sensors

### 4.3 Flight results

MINERVA was deployed on 12 November 2005, when HAYABUSA was very close to the target asteroid Itokawa [24][25] as shown in **Figure 9**. Unfortunately MINERVA did not reach the asteroid surface, because the relative velocity and position of the deployment was not so good. Thus the surface exploration by the rover was not conducted. However MINERVA survived for about 18 hours while the autonomous capabilities worked very well. MINERVA succeeded in taking the picture autonomously and sending the image data via Hayabusa to the earth. **Figure 10** shows the obtained image data, which is a part of solar paddle of Hayabusa spacecraft.

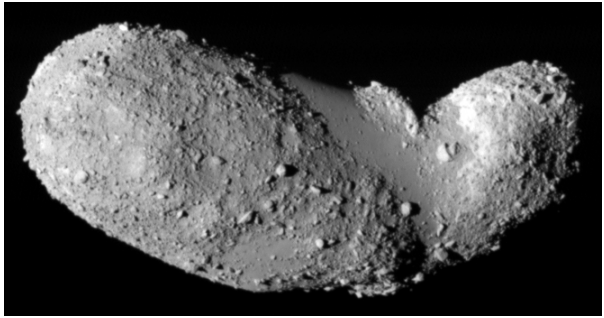

**Figure 9:** Target asteroid Itokawa

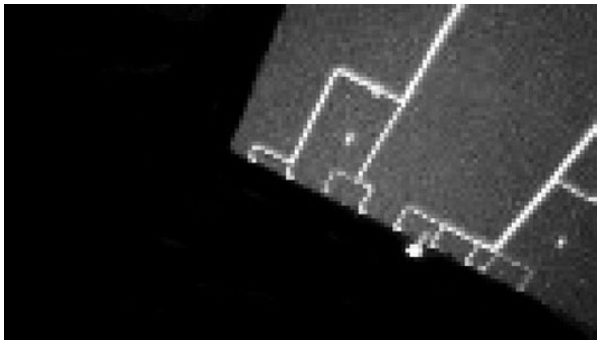

**Figure 10:** Image data taken by MINERVA

### 5 Conclusions

This paper has presented the design of an exploration robot which can move over the surface of small bodies with very weak gravity. Firstly this paper discussed issues related to mobility in a micro-gravity environment. A new mobility system with hopping function was introduced. This paper also described the intelligence of deep space explorer. Then this paper presented the world first asteroid exploration rover "MINERVA", on board the HAYABUSA spacecraft. MINERVA could not explore the surface of the target asteroid Itokawa, but the effectiveness of the most of intelligent functions were confirmed by the flight data.